\documentclass{article}
\usepackage{spconf,amsmath,graphicx}
\usepackage{enumitem}
\setlist{nosep, leftmargin=14pt}

\usepackage{mwe} 
\usepackage{xcolor}
\usepackage{multirow}
\usepackage{etoolbox,lipsum}
\usepackage{graphicx}
\usepackage{caption}

\def\code#1{\texttt{#1}}

\title{Towards Selection of Large Multimodal Models as Engines for Burned-in Protected Health Information Detection in Medical Images}
\name{Tuan Truong \qquad Guillermo Jimenez Perez \qquad Pedro Osorio \qquad Matthias Lenga}
\address{Bayer AG, Berlin, Germany \\ }
\begin{document}
\newcommand{\insertfig}{\includegraphics[width=\textwidth, trim={0cm 8.5cm 0cm 2.5cm}, clip]{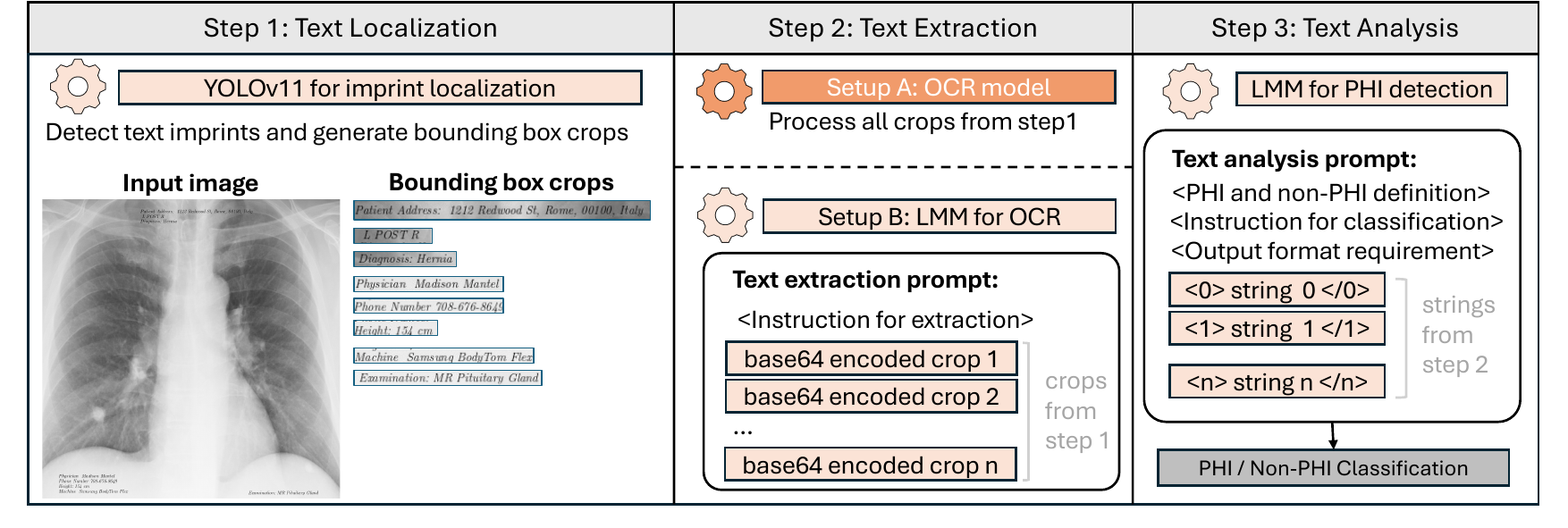}\captionof{figure}{Overview of PHI detection pipeline using Large Multimodal Models (LMMs). Step 1 localizes text regions from medical images. Step 2 compares two text extraction setups: Setup A uses a traditional OCR (EasyOCR) model, while Setup B employs an LMM (GPT-4o, Gemini 2.5 Flash, Qwen2.5-VL 7B) for direct extraction. Step 3 performs PHI vs. non-PHI text classification using an LMM-based prompt.\\}\label{fig:workflow}}
\makeatletter
\apptocmd{\@maketitle}{\centering\insertfig}{}{}
\makeatother
\vspace{-4cm}
\maketitle

\begin{abstract}

The detection of Protected Health Information (PHI) in medical imaging is critical for safeguarding patient privacy and ensuring compliance with regulatory frameworks. Traditional detection methodologies predominantly utilize Optical Character Recognition (OCR) models in conjunction with named entity recognition. However, recent advancements in Large Multimodal Model (LMM) present new opportunities for enhanced text extraction and semantic analysis. In this study, we systematically benchmark three prominent closed and open-sourced LMMs, namely GPT-4o, Gemini 2.5 Flash, and Qwen 2.5 7B, utilizing two distinct pipeline configurations: one dedicated to text analysis alone and another integrating both OCR and semantic analysis. Our results indicate that LMM exhibits superior OCR efficacy (WER: 0.03-0.05, CER: 0.02-0.03) compared to conventional models like EasyOCR. However, this improvement in OCR performance does not consistently correlate with enhanced overall PHI detection accuracy. The strongest performance gains are observed on test cases with complex imprint patterns. In scenarios where text regions are well readable with sufficient contrast, and strong LMMs are employed for text analysis after OCR, different pipeline configurations yield similar results. Furthermore, we provide empirically grounded recommendations for LMM selection tailored to specific operational constraints and propose a deployment strategy that leverages scalable and modular infrastructure.
\end{abstract}

\vspace{-0.02\textwidth}
\section{Introduction}
\label{sec:intro}
In recent years, the detection of Protected Health Information (PHI) has become increasingly important, particularly within the realm of medical data management. Traditional pipelines for PHI detection typically consist of Optical Character Recognition (OCR) followed by analysis steps \cite{vcelak_identification_2019, macdonald_method_2024, monteiro_-identification_2017, rempe_-identification_2024, naddeo_dicom_2025}. However, the advent of Large Multimodal Model (LMM) presents new opportunities for enhancing these processes. While LMM can be utilized in end-to-end pipelines for text analysis, their ability to localize text effectively for subsequent redaction remains a challenge \cite{lee_evaluation_2025, truong_exploring_2025, clunie_summary_2024}. This limitation stems from the fact that general LMMs are not specifically trained for multi-object localization, which increases the risk of incomplete text detection in medical images.

In this paper, we build upon previous research of \cite{lee_evaluation_2025, truong_exploring_2025} by focusing on two specific configurations that leverage LMM effectively. The first configuration utilizes LMM exclusively during the text analysis phase, while the second extends their application to include text extraction. Research by \cite{truong_exploring_2025} has demonstrated that these configurations offer competitive performance; however, integrating LMM into the OCR process significantly raises latency and costs, particularly when using advanced models like GPT-4o \cite{openai_gpt-4o_2024}.

To further study these trade-offs, we introduce two additional models into our benchmarking: Gemini 2.5 Flash \cite{google_deepmind_and_google_research_gemini_2025} and Qwen2.5-VL 7B \cite{qwenlm_qwen25-vl_2025}. Our main contribution is a systematic benchmark of three state-of-the-art LLMs under two PHI detection configurations, revealing key trade-offs between accuracy, latency, and deployability under varying resource constraints and privacy requirements. Specifically, our study addresses the following research questions:

\begin{itemize}
    \item What benefits, if any, can be achieved by replacing traditional OCR models with LMMs to enhance PHI detection?
    \item Which LMM is optimal for selection with respect to performance metrics, latency considerations, and privacy implications?
    \item What key factors should guide the deployment of an LMM-based PHI detection pipeline in real-world healthcare settings?
\end{itemize}

\vspace{-0.02\textwidth}
\section{Methods}
\label{sec:methods}
\vspace{-0.01\textwidth}
In this study, we employ the PHI pipeline put forward by \cite{truong_exploring_2025}, which comprises three integral components: text localization, text extraction, and text analysis (Figure~\ref{fig:workflow}). The text localization module identifies text areas within images, while the text extraction module serves as an OCR engine, converting pixel-level text into machine-encoded text. The final component, text analysis, focuses on identifying PHI and related entities from the extracted text.

To assess the performance of these models, we establish two distinct evaluation setups. In the first setup, each LMM operates exclusively as the text analysis module. In the second setup, the LMMs also function as the OCR engine, extracting text from localized image crops. This structured evaluation allows us to comprehensively analyze each model's capabilities and their suitability for specific tasks within the PHI pipeline.

To refine our candidate list of LMMs, we conducted a preliminary experiment to evaluate OCR capabilities of different LMMs and dedicated OCR models on medical image crops from RadPHI-test dataset (see Section~\ref{sec:experiments-data}).  The RadPHI-test dataset was chosen due to its diverse imprint variations, which simulate real-world challenges in PHI detection. Apart from GPT-4o, which is a flagship model and was included in previous benchmarks \cite{lee_evaluation_2025, truong_exploring_2025}, the other models in the list were selected based on their suitability for specific operational needs. These include dedicated OCR models and lighter LMMs that prioritize low inference latency and cost-efficiency. The results, summarized in Table \ref{tab:ocr-benchmarking}, are characterized by word error rate (WER) and character error rate (CER). We subsequently shortlisted the top three models based on their performance across various aspects of our requirements.

The first model, GPT-4o, emerges as a robust foundational model, achieving superior performance across a range of tasks. The second model, Gemini 2.5 Flash, is notable for its slower inference latency and lower operational costs while maintaining competitive performance compared to other leading proprietary models. This balance makes it a vital consideration in the selection of LMMs, especially for processes involving large batches of data. Lastly, the third model, Qwen2.5-VL 7B, is a lightweight, open-source option. Its ability to efficiently run a compact model is particularly advantageous in scenarios where data privacy is critical, such as in medical applications, as it is more cost-effective and easier to deploy than larger models.

\begin{table}[ht!]
\centering
\resizebox{0.6\columnwidth}{!}{%
\begin{tabular}{|l|c|c|}
\hline
\textbf{Model name} & \textbf{WER $\downarrow$} & \textbf{CER $\downarrow$} \\
\hline
GPT-4o \cite{openai_gpt-4o_2024}          & 0.03 & 0.02 \\
Gemini-2.5-flash \cite{google_deepmind_and_google_research_gemini_2025} & 0.05 & 0.03 \\
Qwen2.5-VL 7B \cite{qwenlm_qwen25-vl_2025}      & 0.05 & 0.03 \\
InternVL2.5 8B \cite{chen_expanding_2025}  & 0.09 & 0.10 \\
TrOCR \cite{li_trocr_2021}           & 0.13 & 0.05 \\
Qwen2.5-VL 3B  \cite{qwenlm_qwen25-vl_2025}     & 0.13 & 0.06 \\
GOT-OCR \cite{wei_general_2024}         & 0.17 & 0.06 \\
InternVL2.5 4B \cite{chen_expanding_2025} & 0.26 & 0.59 \\
EasyOCR \cite{noauthor_jaidedaieasyocr_nodate}         & 0.29 & 0.07 \\
\hline
\end{tabular}
}
\caption{Performance comparison of different models on the OCR task using the RadPHI-test dataset, ranked by WER.}
\label{tab:ocr-benchmarking}
\end{table}
\vspace{-0.05\textwidth}
\section{Experiments}
\label{sec:experiments}
\vspace{-0.02\textwidth}
\subsection{Data}
\label{sec:experiments-data}

\begin{figure}
    \centering
    \includegraphics[width=0.4\textwidth, trim={0cm 11cm 15cm 2cm}, clip]{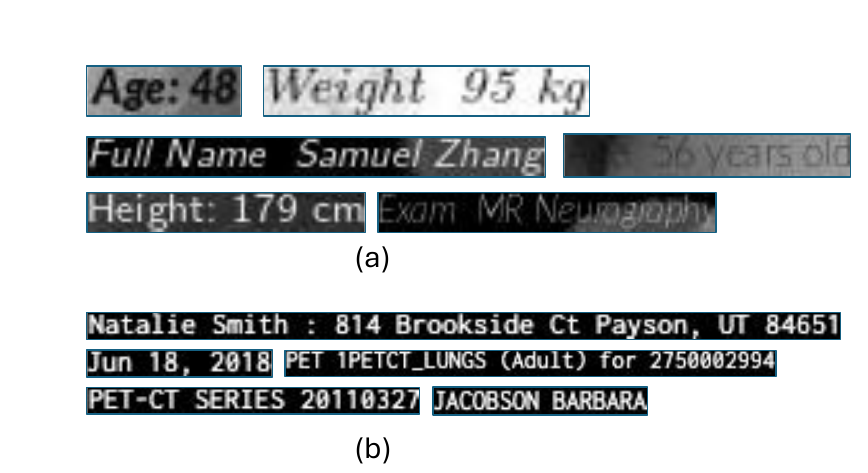}
    \caption{Examples of imprints from (a) RadPHI-test and (b) MIDI datasets.}
    \label{fig:data}
    \vspace{-0.04\textwidth}
\end{figure}

We use two datasets \textbf{RadPHI-test} and \textbf{MIDI} which were introduced in \cite{truong_exploring_2025} and are publicly available\footnote{https://doi.org/10.5281/zenodo.17201610}. These datasets cover a range of imaging modalities, imprint variations, and text complexities to support robust benchmarking.
RadPHI-test (Figure~\ref{fig:data}a) includes 1000 radiological images equally distributed across four modalities with synthetic imprints generated at different locations, fonts, sizes, and contrasts to background. The MIDI dataset (Figure~\ref{fig:data}b) is derived from the 2024 MIDI-B challenge \cite{rutherford_medical_2025} and includes images overlaid with DICOM header metadata using a standard DICOM viewer export. All images are labeled with bounding boxes and imprint categories. While MIDI presents challenges due to the presence of redundant PHI elements, like repeated dates and identifiers, the imprint variations are less pronounced compared to a the purely synthetic and heavily randomized imprints from RadPHI-test. The text categories that are considered PHI include patient names, addresses, identifiers, dates, phone numbers, and emails. Each image in RadPHI-test can have maximum eight imprints, whereas it is up to 40 imprints in MIDI. Table~\ref{tab:datasets} provides a summary of PHI statistics of both datasets.

\begin{table}[ht!]
    \centering
    \resizebox{\columnwidth}{!}{%
    \begin{tabular}{|l|c|c|c|}
        \hline
        \textbf{Dataset} & \textbf{No. Images} & \textbf{No. PHI Images} & \textbf{No. PHI imprints }\\
        \hline
        RadPHI-test & 1000 & 779 & 1550 \\
        \hline
        MIDI & 550 & 426 & 1428 \\
        \hline
    \end{tabular}
    }
    \caption{Overview of RadPHI-test and MIDI datasets}
    \label{tab:datasets}
\end{table}

\vspace{-0.05\textwidth}
\subsection{Experimental Setup}
\label{sec:experimental-setup}
Following \cite{truong_exploring_2025}, we are benchmarking two LMM-based setups. Both setups utilize the YOLOv11 \cite{ultralytics_ultralytics_2024} architecture for imprint localization. \textbf{Setup A} employs EasyOCR for text extraction, followed by subsequent text analysis performed by LMM. In contrast, \textbf{Setup B} integrates LMM for both text extraction and analysis. In both setups, images are processed sequentially at each stage of the workflow, meaning that they are not processed in batches. Once all images have been processed at a given stage, they advance to the subsequent stage for further handling.

For text extraction, Setup A utilizes EasyOCR to sequentially infer text from each individual image crop. In setup B, the prompts provided to the LMM are designed to be straightforward, containing clear instructions to extract text from specified image crops. To optimize processing time and reduce token usage, crops are aggregated into chunks for API calls, with the size of 10 crops per call for GPT-4o and Gemini 2.5 Flash, while Qwen2.5-VL 7B is limited to five crops due to a shorter context window.

For the text analysis phase, the input to the LMM is structured as a list of strings, each tagged with a unique identifier, for example, \code{["<0> Patient name: John Doe </0>", "<1> Age: 24 </1>"]}. This tagging approach enables the system to reference multiple PHI terms within a single bounding box.The prompt structure for text analysis is systematically organized to include: 1) an introduction defining the distinctions between PHI and non-PHI content, 2) explicit instructions for identifying potential OCR errors and classifying the content accurately, and 3) requirements for output as a JSON object containing classification and rationale. This standardized prompt structure is consistently applied across all LMMs utilized within both setups.

\subsection{Metrics}
\label{sec:experiments-eval}
The classification performance for PHI is quantified using micro-precision and recall at the instance level. A true positive instance is defined as one in which the PHI is not only accurately identified but also correctly matched to the corresponding text coordinates.
In addition to these accuracy metrics, we also measure the average inference latency per image (in seconds). 
This latency is calculated by dividing the pipeline's processing time over the dataset by the total number of images. The text localization model is implemented on a Tesla V100 GPU with 16 GB of GPU memory. Proprietary models, such as GPT-4o and Gemini 2.5 Flash, are accessed via API endpoints, while Qwen2.5-VL 7B is deployed locally on the same Tesla V100 GPU using the Ollama framework\footnote{https://ollama.com/}. To account for the stochastic nature of LMM, we conduct five runs per setup and model, reporting the average metrics.

\section{Results}
\label{sec:results}

\subsection{RadPHI-test}
Table~\ref{tab:results-radphi} presents a summary of the performance and latency metrics for three benchmark LMMs evaluated on the RadPHI-test dataset. Among the models, GPT-4o achieves the highest overall precision and recall, both exceeding 0.99, followed by Gemini 2.5 Flash and Qwen 2.5-VL 7B. The performance of the models is generally higher when employing Setup B, which uses an LMM for text extraction, compared to Setup A. An exception to this trend is observed with GPT-4o, which shows slightly higher precision in Setup A than in Setup B. Regarding latency, GPT-4o exhibits the highest latency levels, with Qwen 2.5-VL and Gemini 2.5 Flash following. Transitioning from EasyOCR (Setup A) to the language model (Setup B) results in notable increases in latency: 19\% for GPT-4o, 35\% for Gemini 2.5 Flash, and 38\% for Qwen 2.5-VL 7B.

\subsection{MIDI}
As shown in Table~\ref{tab:results-midi}, GPT-4o and Gemini 2.5 Flash exhibit competitive performance in both precision and recall, followed by  Qwen2.5-VL 7B. It is observed that Setup A achieves better overall precision and recall compared to Setup B, suggesting that on this benchmark the integration of LMM for text extraction does not enhance the final PHI classification results.

Transitioning from Setup A to Setup B results in latency increases of 33\%, 48\%, and 76\% for GPT-4o, Gemini 2.5 Flash, and Qwen2.5-VL 7B, respectively. In comparison to the RadPHI-test, the ranking of performance remains consistent; however, both GPT-4o and Gemini 2.5 Flash display slight reductions in precision and recall, while Qwen2.5-VL 7B experiences a more significant decline.

The notable increase in latency between RadPHI-test and MIDI can be attributed to the greater number of imprints per image in the MIDI dataset, which requires additional processing time for text extraction and analysis. Although the Qwen model performs comparably to other models in the RadPHI-test, it underperforms in the MIDI context due to challenges in processing a large volume of text per call and difficulties in accurately interpreting context when classifying PHI. For example, it struggles to differentiate between study and patient identifiers, with the former not qualifying as PHI. In addition, it often classifies placeholders such as "PATIENT NAME" or "IDENTIFIER" without actual PHI values as PHI, leading to lower precision compared to the other counterparts. This illustrates the reasoning limitations that lighter-weight models may encounter when addressing complex and nuanced cases.


%
\begin{table}[ht!]
\centering
\resizebox{\columnwidth}{!}{%
\begin{tabular}{lcccc}
\hline
\textbf{LMM} & \textbf{Setup} & \textbf{Precision} $\uparrow$ & \textbf{Recall} $\uparrow$ &  \textbf{Latency} $\downarrow$\\
\hline
GPT-4o & A & \textbf{0.994} & 0.988 & \textbf{4.143}\\
GPT-4o & B &  0.992 & \textbf{0.993} & 4.920\\
\hline
Gemini 2.5 Flash & A & 0.955 & 0.967 & \textbf{2.651}\\
Gemini 2.5 Flash & B & \textbf{0.973} & \textbf{0.972} & 3.588\\
\hline
Qwen2.5-VL 7B & A & 0.962 & 0.943 & \textbf{3.070}\\
Qwen2.5-VL 7B & B & \textbf{0.964 }& \textbf{0.953} & 4.233\\
\hline
\end{tabular}}
\caption{Comparison of LMM performance and latency on the \textbf{RadPHI-test} dataset under OCR-based (Setup A) and LMM-based (Setup B) text extraction.}
\label{tab:results-radphi}
\end{table}

\begin{table}[ht!]
\centering
\resizebox{\columnwidth}{!}{%
\begin{tabular}{lcccc}
\hline
\textbf{LMM} & \textbf{Setup} & \textbf{Precision} $\uparrow$ & \textbf{Recall} $\uparrow$  &  \textbf{Latency} $\downarrow$\\
\hline
GPT-4o & A & \textbf{0.981} & \textbf{0.985} & \textbf{13.679}\\
GPT-4o & B & 0.979 & 0.977 & 18.110\\
\hline
Gemini 2.5 Flash & A & \textbf{0.967} & \textbf{0.988} & \textbf{5.937}\\
Gemini 2.5 Flash & B & 0.963 & 0.985 & 8.775\\
\hline
Qwen2.5-VL 7B & A & 0.719 & \textbf{0.858} & \textbf{8.011}\\
Qwen2.5-VL 7B & B & \textbf{0.751} & 0.843 & 14.099\\
\hline
\end{tabular}}
\caption{Comparison of LMM performance and latency on the \textbf{MIDI} dataset under OCR-based (Setup A) and LMM-based (Setup B) text extraction.}
\label{tab:results-midi}
\end{table}

\subsection{Failure Analysis}
\label{sec:results-failure-analysis}
While setup B exhibits a marginal improvement in performance on the RadPHI-test dataset, results on the MIDI dataset show the opposite trend. A manual inspection of predictions reveals variations linked to dataset characteristics.

First, both GPT-4o and Gemini 2.5 Flash struggle with extremely small images (under 12 pixels), impairing their ability to recognize critical characters. This challenge is more evident in the MIDI dataset, where misinterpretations, such as confusing '0' with '3', '6', or '8', can lead to inaccurate transcriptions like '20160730' being rendered as '28168730'. Such errors can cause misclassification as non-PHI values, particularly when dates appear as study or series IDs, making generic placeholders ineffective.

Second, occasional API call errors, resulting from traffic issues, hallucinations, or incompatibility with Pydantic parsing, are more frequent in LMM-based text extraction with image inputs, especially in the MIDI dataset with its higher volume of imprints. These issues result in missing text extractions and classifications, which contribute to the slightly lower performance of LMM-based text extraction compared to traditional OCR methods in some cases.



\section{Discussion}
\label{sec:discussion}

\subsection{LMM for Text Extraction}
The evaluation of OCR capabilities indicates that LMMs generally outperform traditional OCR models (Table~\ref{tab:ocr-benchmarking}), although trade-offs depend on the dataset. On the RadPHI-test dataset, featuring complex and partially obscured imprints, the LMM-based approach (Setup B) shows slight improvements but incurs a 20\% to 40\% increase in inference latency. In contrast, the dedicated OCR pipeline (Setup A) achieves comparable or slightly superior accuracy on the simpler, more legible MIDI dataset while reducing inference latency by up to 70\%. These results suggest that a dedicated OCR model is more efficient for clear imprints, while LMM excels in challenging visibility conditions.

Further insights from the failure analysis in Section~\ref{sec:results-failure-analysis} indicate that specific issues may be mitigated through engineering solutions. For example, implementing error-catching mechanisms for API calls may enhance reliability. Additionally, asynchronous API calls could significantly reduce latency. Exploring a hybrid approach, where less confident traditional OCR predictions are verified by an LMM, might balance latency reduction with improved classification robustness.

\subsection{LMM Selection for PHI Pipeline}
Selecting an appropriate language model for medical applications such as PHI detection requires a balanced evaluation of accuracy, latency, and data privacy considerations. While the OCR capabilities of the evaluated models are broadly comparable, their text reasoning performance and inference efficiency differ substantially. 

Proprietary models, such as GPT-4o, deliver the highest precision and recall but exhibit significantly higher inference latency compared to lighter alternatives, limiting their suitability for real-time workflows, though they remain well-suited for batch or offline analyses where maximum PHI detection accuracy is prioritized. Therefore, if the use case involves batch or offline prediction and needs to ensure the completeness of PHI detection, GPT-4o is the recommended option.

Within proprietary model families, lighter variants, such as Gemini 2.5 Flash, strike a favorable trade-off, achieving competitive accuracy with substantially lower latency. This makes them appealing for time-sensitive tasks, such as reviewing de-identified images where real-time feedback is required. However, their closed-source nature and cloud-dependent deployment may introduce privacy and compliance concerns in regulated healthcare environments. 

In contrast, open-source models, like Qwen2.5-VL 7B, offer greater flexibility for on-premise deployment and fine-tuning, allowing adaptation to institution-specific data and privacy requirements. While their out-of-the-box performance is lower compared to larger proprietary models, targeted fine-tuning or instruction adaptation can potentially bridge this gap. This makes them a practical and privacy-preserving alternative for PHI detection pipelines. In scenarios where privacy is a dominant concern and local deployment is mandatory, we recommend using an open-source model such as Qwen2.5-VL 7B with proper fine-tuning.

\subsection{Deployment Considerations}

The appropriate deployment infrastructure to serve the PHI pipeline should be selected based on the specific needs of the target PHI workflow, as well as considerations regarding privacy and compliance. For instance, cloud based deployments are usually the preferred approach as they offer easy scalability, a fully managed environment, and reduced infrastructure management overhead. However, on-premises deployments could be chosen in scenarios when organizations require full data control for strict compliance requirements.
For our deployment we selected the Google Cloud Platform (GCP) with additional security layers provided by the Bayer AI Innovation Platform (AIIP) \cite{bayer_ag_ai_2025} to ensure that all data resides in an secure, isolated environment which is HIPAA and GDPR compliant. While this study utilized GCP services for deployment, similar technologies exist from other vendors, e.g., Microsoft Azure or Amazon Web Services, and could be chosen depending on organizational preferences and constraints.

In workflows where real-time feedback is required, such as for live PHI redaction in the radiologist workstation, the pipeline needs to be served on a stable and responsive endpoint with low latency responses. This was achieved in GCP by using a combination of services: a proxy Cloud Run \cite{google_cloud_cloud_2025} service, which handles the requests from the live redaction front end; an inference GPU bound Cloud Run \cite{google_cloud_cloud_2025} service which receives requests from the proxy triggering prediction jobs concurrently; and, finally, a Cloud SQL \cite{google_cloud_cloud_nodate} database which tracks job statuses and stores predictions from the inference service to be retrieved to the front end through the proxy. This modular design using Cloud Run allows independent serverless auto-scaling of both the client-facing proxy service and the inference service, thus handling varying workloads efficiently.
In fact, our preliminary benchmarking shows that, if we chunk the RadPHI dataset into 100 separate detection requests, the deployed PHI detection pipeline (Setup B with Gemini 2.5 Flash) is able to process it entirely in a total of 128 seconds (0.13 seconds per image) with a mean response time (latency) of 16.5 ± 4.1 seconds per request (1.7 ± 0.4 seconds per image using batching).
For higher workloads, such as 1000 simultaneous PHI detection requests (10000 images), the infrastructure automatically scales up to handle this high demand in parallel, thus returning all responses in only 153 seconds (0.02 seconds per image).
On the other hand, for workflows that do not require live redaction, the pipeline could be deployed using offline prediction services, such as GCP's Vertex AI batch prediction \cite{google_cloud_vertex_nodate}, enabling batch de-identification before data sharing.






\section{Conclusion}
\label{sec:conclusion}

In this study, we analyze the strengths and limitations of state-of-the-art models like GPT-4o, Gemini 2.5 Flash, and Qwen2.5-VL 7B under two different configurations of the LMM-based PHI detection pipeline. The paper emphasizes the nuanced trade-offs between accuracy, latency, and privacy, providing practical recommendations for deployment in healthcare settings. The findings underscore the potential of LMM as a powerful tool for PHI detection but also highlight the need for hybrid or optimized solutions to mitigate its technical and computational drawbacks. This benchmarking effort serves as a valuable reference for researchers and practitioners working on de-identification pipelines in the medical domain. Future work should focus on validating these findings in real-world clinical environments, exploring fine-tuning strategies for open-source models, and implementing latency-aware hybrid approaches that balance efficiency and reliability.



\section{Compliance with ethical standards}
\label{sec:ethics}
This research study was conducted retrospectively using human subject data derived from subsets of open-access data, which was introduced and published in the work of \cite{truong_exploring_2025}. Ethical approval was not required, as confirmed by the license attached to the sub-datasets.

\section{Acknowledgments}
\label{sec:acknowledgments}
The authors would like to thank the Bayer team of the AI Innovation Platform for providing computing infrastructure and technical support.

\bibliographystyle{IEEEbib}
\bibliography{refs}

\end{document}